\def\year{2021}\relax
\documentclass[letterpaper]{article} % DO NOT CHANGE THIS
\usepackage{aaai21}  % DO NOT CHANGE THIS
\usepackage{times}  % DO NOT CHANGE THIS
\usepackage{helvet} % DO NOT CHANGE THIS
\usepackage{courier}  % DO NOT CHANGE THIS
\usepackage[hyphens]{url}  % DO NOT CHANGE THIS
\usepackage{graphicx} % DO NOT CHANGE THIS
\usepackage{natbib}  % DO NOT CHANGE THIS AND DO NOT ADD ANY OPTIONS TO IT
\usepackage{caption} % DO NOT CHANGE THIS AND DO NOT ADD ANY OPTIONS TO IT
\usepackage{subcaption}
\usepackage{soul}
\usepackage{amsmath, amsthm, amssymb}
\usepackage{algorithm,algorithmic,paralist}
\usepackage{bm}
\usepackage[american]{babel}
\usepackage{microtype}
\usepackage{latexsym}
\usepackage{booktabs}
\usepackage{multirow}
\usepackage{array}
\usepackage{thmtools}
\usepackage{thm-restate}
\usepackage{cleveref}
\usepackage{amsthm}
\usepackage{multicol}
\usepackage{multirow}
\newtheorem{theorem}{Theorem}

\newtheorem{corollary}{Corollary}[theorem]
\usepackage{mathtools}

\title{Improving Model Robustness by Adaptively Correcting \\Perturbation Levels with Active Queries}
\author{
    Kun-Peng Ning\thanks{Equal contribution.},
    Lue Tao\footnotemark[1],
    Songcan Chen,
    Sheng-Jun Huang\thanks{Correspondence to: Sheng-Jun Huang $($huangsj@nuaa.edu.cn$)$.}\\
}
\affiliations{
    College of Computer Science and Technology, Nanjing University of Aeronautics and Astronautics \\
    MIIT Key Laboratory of Pattern Analysis and Machine Intelligence \\
    Collaborative Innovation Center of Novel Software Technology and Industrialization, Nanjing, 211106 \\
}
\begin{document}

\maketitle

\begin{abstract}
In addition to high accuracy, robustness is becoming increasingly important for machine learning models in various applications. Recently, much research has been devoted to improving the model robustness by training with noise perturbations. Most existing studies assume a fixed perturbation level for all training examples, which however hardly holds in real tasks. In fact, excessive perturbations may destroy the discriminative content of an example, while deficient perturbations may fail to provide helpful information for improving the robustness. Motivated by this observation, we propose to adaptively adjust the perturbation levels for each example in the training process. Specifically, a novel active learning framework is proposed to allow the model to interactively query the correct perturbation level from human experts. By designing a cost-effective sampling strategy along with a new query type, the robustness can be significantly improved with a few queries. Both theoretical analysis and experimental studies validate the effectiveness of the proposed approach.
\end{abstract}

\section{Introduction}

\begin{figure*}[t]
	\begin{center}
		\includegraphics*[width=0.995\textwidth]{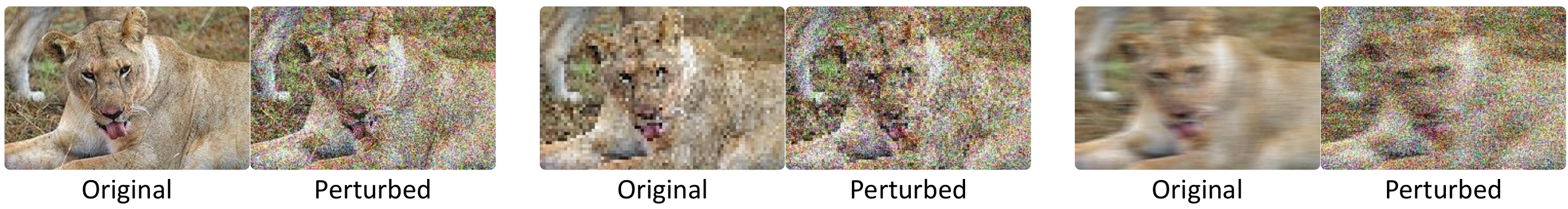}
		\caption{The influence of the same perturbation ($\mathcal{N}(0, \sigma^2 I)$, where $\sigma=0.23$) on images with different intrinsic robustness. The three perturbed images are generated by corrupting the three original images respectively with the same perturbation.}
		\label{fig:example}
	\end{center}
\end{figure*}

%介绍DNNs的输出受扰动影响。   需要介绍我们的框架适用于更大的场景，例如对抗噪声、高斯噪声以及一些其它的噪声都行，但是我们本文只考虑高斯噪声。
Deep Neural Networks (DNNs) have achieved great success in many tasks with high accuracy ~\cite{krizhevsky2012imagenet, he2015delving, sutskever2014sequence,silver2017mastering}. On the other hand, deep models are less robust when applied to datasets with noise perturbations~\cite{szegedy2013intriguing, alzantot2018generating, hendrycks2019robustness}. Much research has been devoted to mitigating this issue in recent years~\cite{geirhos2018imagenet, rusak2020increasing, tramer2020adaptive, hendrycks2019using, mao2019metric, hendrycks2019augmix, zhang2019making}. Roughly speaking, existing studies are trying to improve the model robustness by handling two different categories of perturbations. One is \textit{adversarial perturbations}, which are maliciously designed to fool the models under some distance constraint (e.g. $\ell_{\infty}$ distance~\cite{madry2017towards} or Wasserstein distance~\cite{wong2019wasserstein}), while the appearance contents are preserved. The other one is \textit{corruption perturbations}, which are usually incidentally generated during the process of data collection and editing (e.g. Gaussian noise \cite{chapelle2001vicinal}, motion blur~\cite{hendrycks2019robustness}).

In this paper, we focus on the latter case and try to handle corruption perturbations for improving the model robustness. Corruption perturbation problem is becoming a ubiquitous challenge in various applications \cite{hendrycks2019robustness, michaelis2019benchmarking}. 
% Constantly increasing number of deep learning based systems trained and evaluated in laboratory environments have been deployed in real-world applications. Extensive and unexpected noises exist in the input data collected from these real environments, and may cause serious failure if the models are not robust enough. 
More and more systems based on deep learning have been deployed to real-world applications. They are typically trained and evaluated in laboratory environments. However, extensive and unexpected noises exist in real environments, which may cause serious failures if the models are not robust enough.
For example, autonomous vehicles need to be able to cope with wildly varying outdoor conditions such as fog, frost, snow, sand storms, or falling leaves~\cite{michaelis2019benchmarking}. Likewise, speech recognition systems should perform well regardless of the additive noise or convolutional distortions~\cite{qian2016very}.

Training DNNs on perturbed examples is the primal approach to improve the model robustness \cite{carlini2019evaluating, hendrycks2019augmix}. Representative methods include noise injection \cite{grandvalet1997noise} and PGD-based robust training \cite{madry2017towards}. However, most of the existing methods assign a fixed level of perturbations (e.g. fixed radius in $\ell_p$ norm-bounded perturbations or bandwidth in Gaussian noise) to all examples, ignoring the fact that each example has its own intrinsic tolerance to noises. In fact, excessive perturbations would destroy the class-distinguishing feature of an example, while deficient perturbations may fail to provide helpful information for improving the robustness. Intuitively, some examples are closer to the decision boundary, where tiny perturbations could change their labels, while some others are far away from the decision boundary and may tolerate higher levels of perturbation. As shown in Figure~\ref{fig:example}, under the same perturbation, the discriminability of the corrupted images is significantly different, if the original images have different intrinsic robustness. A higher-quality image is likely to be able to tolerate heavier perturbations.

Several recent works in the literature seek to adjust the perturbation levels for different examples according to prediction loss~\cite{cheng2020cat, sitawarin2020improving, zhang2020attacks}. While it is intuitively reasonable to assign higher perturbations to examples with smaller losses, these methods may suffer from model bias and lack a correction strategy to seek help from ground-truth supervision. In the case that the model is unreliable, the perturbation adjustment strategy may be seriously misled to hurt model performance. It is thus rather important to allow the learning system to query the ground-truth information about the perturbation levels for robustness. Such an idea has been widely used in other machine learning tasks. For example, in the active learning literature, learning algorithms are allowed to query class labels from human experts to improve model accuracy \cite{settles2009active}. Actually, the human recognition system is remarkably robust against a wide range of noises and corruptions~\cite{rusak2020increasing}, and thus can identify the proper tolerance level of perturbations for a specific image. This motivates us to query the perturbation levels from human experts to train a robust model.

In this paper, we propose to adaptively adjust the perturbation levels for each training example, along with a querying strategy to get ground-truth information from the human experts to correct the perturbation levels. It is worth to note that the human annotation could be costly, and thus it is less practical to query the perturbation levels for all examples. To overcome this challenge, we propose a novel active learning framework to \textbf{A}ctively \textbf{Q}uery \textbf{P}erturbation \textbf{L}evels (\textbf{AQPL} for short), aiming to train a robust model with least queries. Specifically, at each iteration of active learning, we first estimate the conformity of the current perturbation level for each example based on the prediction consistency over multiple generated noises, and then actively select the examples with the least conformity for querying. In this way, the examples with over large or over small perturbations will be corrected with the queried ground-truth information. To further reduce the annotation cost, a cost-effective query type is designed to allow human experts to easily decide the perturbation level for an image.

Experiments are conducted on multiple datasets with variant noise perturbations. Our results validate the effectiveness of the proposed AQPL method with adaptive correction of perturbation levels. Model robustness, as well as accuracy, are significantly improved by actively querying a very few times with low cost.

The main contributions of this work are summarized as follows.
\begin{itemize}
	\item A novel framework AQPL is proposed to improve model robustness via querying the perturbation level of examples. It is a new attempt to improve the model robustness by interacting with human experts.
	
	\item An effective strategy is proposed to actively select the most useful example for perturbation level correction, which significantly reduces the query numbers for robust training.
	
	\item A cost-effective query type is designed to allow human experts to easily decide the proper perturbation level of an image with low annotation cost.
\end{itemize}

The rest of the paper is organized as follows. We review related work in Section 2 and introduce the proposed method in Section 3. Section 4 reports the experiments, followed by the conclusion in Section 5.

\begin{figure*}[t]
	\centering
	\includegraphics*[width=\textwidth]{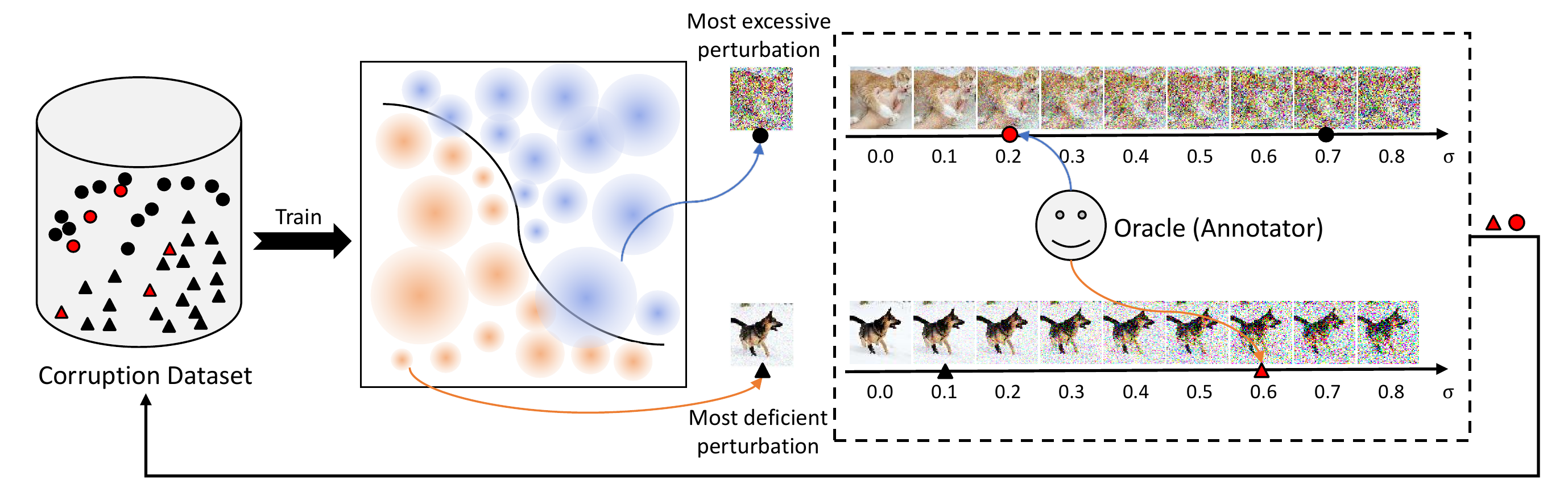}
	\caption{The proposed AQPL framework. 
		Based on the model trained on the corruption dataset, two examples with most excessive and most deficient perturbations are selected for querying. Then, the perturbation levels (indicated by black marker) are corrected to the proper perturbation levels (indicated by red marker) by the annotator. After that, the corruption dataset and the model are updated.
	}
	\label{fig:AQPL}
\end{figure*}

\section{Related Work}

\textbf{Model robustness.}\quad Improving model robustness refers to the goal of ensuring the performance of machine learning models under a variety of imperfect testing conditions, which has a long history. Double backpropagation algorithm~\cite{drucker1991double} is one of the earliest attempt to make models resistant to local minimal perturbations by regularizing input gradients. It reaches a consensus with Tikhonov regularization~\cite{tikhonov1977solutions}, which is equivalent to training with noise under the assumption that the noise amplitude is small enough~\cite{bishop1995training}. Nevertheless, a more practical and ubiquitous situation in the real-world is some nonnegligible noises that exist in input data that we can easily tell, such as fog, rain, or falling leaves in camera data under varying outdoor conditions. Such a situation considering perturbations in input data is known as vicinal risk minimization~\cite{chapelle2001vicinal}, along with adversarial risk minimization~\cite{uesato2018adversarial}, sparks a surge of interest in model robustness recently.
To this end, a number of methods have been proposed to mitigate the problem and improve the robustness of DNNs, among which finally training with perturbations remains the most effective one~\cite{carlini2019evaluating, hendrycks2019augmix, rusak2020increasing}.
To achieve this goal, current methods assume a fixed perturbation level for all training examples \cite{madry2017towards, wang2020improving, zhang2019theoretically, rusak2020increasing, Cemgil2020Adversarially}, which hardly holds in real tasks. Most recently, \cite{cheng2020cat, sitawarin2020improving, zhang2020attacks} propose to adaptively adjust the perturbation level for each example according to the capacity of the model-on-training. They adjust the perturbation level of each example to cater to the model-on-training, while we choose to define the intrinsic robustness of examples by quering the oracle.
Since model vulnerability can be view as a purely human-centric phenomenon, and achieving models that are robust and interpretable will require explicitly encoding human priors into the training process~\cite{ilyas2019adversarial}. In this paper, while we focus on improving model robustness to vicinal risk by adaptively correcting perturbation levels with active queries, our active learning framework can also be extended to tackle the adversarial risk minimization problem.

\textbf{Active learning.}\quad Active learning has achieved a great success for learning with limited labeled data. Most researches focus on designing effective sampling strategies to make sure that the selected examples can improve the model performance most~\cite{fu2013survey}. During the past decades, many criteria have been proposed for selecting examples~\cite{fu2013survey,huang2010active,lewis1994sequential,seung1992query,you2014diverse,geman1992neural,roy2001toward}. Among of these approaches, some of them prefer to select the most informative examples to reduce the model uncertainty~\cite{lewis1994sequential,seung1992query,you2014diverse}, while some others prefer to select the most representative examples to match the data distribution~\cite{geman1992neural,roy2001toward}. Moreover, some studies try to combine informativeness and representativeness to achieve better performance~\cite{huang2013active,huang2010active}. Standard active learning methods often ask the oracle to annotate data examples~\cite{fu2013survey}, ~\cite{huijser2017active} tries to improve the classification model by asking for annotations of decision boundary. Similarly, our approach attempts to improve the model robustness by querying for annotations of perturbation level.

\section{The Proposed Approach}
In this section, we first formalize the framework for improving model robustness via active querying, and then introduce the proposed AQPL approach in detail, followed by the theoretical analysis on the active selection strategy.

\subsection{Problem Setting}
%介绍问题设定、动机
We denote by $\mathcal{D}$ the clean dataset with $n$ examples, i.e., $\mathcal{D}=\{(\mathbf{x}_1,y_1),(\mathbf{x}_2,y_2),...,(\mathbf{x}_n,y_n)\}$, where $\mathbf{x}_i \in \mathbb{R}^d$ is the feature vector and $y_i \in \{1,...,K\} =: \mathcal{Y}$ is the ground-truth label. We also denote by $\mathcal{C}$ the dataset with common corruptions (e.g. ImageNet-C), i.e., $\mathcal{C}=\{(\hat{\mathbf{x}}_1,y_1),(\hat{\mathbf{x}}_2,y_2),...,(\hat{\mathbf{x}}_n,y_n)\}$, where $\hat{\mathbf{x}}_i$ is the corrupted instance of $\mathbf{x}_i$ with perturbations.

A model $F_\theta(\mathbf{x}): \mathbb{R}^d \rightarrow \mathcal{Y}$ parameterized by $\theta$ can be trained with the clean dataset $\mathcal{D}$, which however is usually less robust when applied to $\mathcal{C}$ due to the unseen corruptions. To address this issue, the mainstream methods try to train models with noise to improve the robustness against corruption perturbations. Formally, as done in \cite{chapelle2001vicinal, gilmer2019adversarial, rusak2020increasing}, we can improve the classifier $F_\theta$ by minimizing the cross-entropy loss $\mathcal{L}$ on clean dataset $\mathcal{D}$ with additive noise:
\begin{equation}\label{EQ1}
\min _{\theta} \sum_{i=1}^{n}\left[\mathbb{E}_{\boldsymbol{\varepsilon} \sim \mathcal{P}(\sigma)}\left[\mathcal{L}\left(F_{\theta}\left(\mathbf{x}_{i}+\boldsymbol{\varepsilon}\right), y_{i}\right)\right]\right],
\end{equation}
where $\boldsymbol{\varepsilon}$ is the random noise generated according to the noise distribution $\mathcal{P}(\sigma)$, and $\sigma$ is the perturbation level controlling the intensity of the noise. Here $\mathcal{P}(\sigma)$ can be any general noise distribution. Obviously, by minimizing the loss function, the classifier $F_\theta$ will be optimized to correctly recognize the examples perturbed by noise. In previous methods~\cite{madry2017towards, rusak2020increasing, gilmer2019adversarial}, $\sigma$ is either kept fixed or chosen uniformly from a fixed set of standard deviations. However, as discussion above, it is impractical to set a global constant $\sigma$ for all examples, because each example has its own intrinsic robustness towards noises. Therefore, in this paper, we propose a more practical setting where each example has its own perturbation level. Formally, we introduce instance-dependent perturbation level $\sigma_i$ to generate noises for each $x_i$, and define a new loss function as follows:
\begin{equation}\label{EQ2}
\min _{\theta} \sum_{i=1}^{n}\left[\mathbb{E}_{\boldsymbol{\varepsilon}_{i} \sim \mathcal{P}\left(\sigma_{i}\right)}\left[\mathcal{L}\left(F_{\theta}\left(\mathbf{x}_{i}+\boldsymbol{\varepsilon}_{i}\right), y_{i}\right)\right]\right].
\end{equation}
Obviously, image with stronger intrinsic robustness should receive higher values of $\sigma_i$, while image with weaker intrinsic robustness should receive lower values of $\sigma_i$. While the initialized perturbation levels are likely not to conform with the intrinsic robustness, we actively select the most useful examples and query their ground-truth information to adaptively correct the perturbation levels.

\subsection{Algorithm Detail}
%介绍和传统的主动学习的区别

The proposed framework AQPL is demonstrated in Figure~\ref{fig:AQPL}. Firstly, all examples are assigned an initial perturbation level. Then at each iteration, based on the proposed conformity criterion, the two most useful examples (one with most excessive perturbation, and one with most deficient perturbation) are selected for perturbation level correction. After that, the oracle is asked to annotate a proper perturbation level that conform with the intrinsic robustness for the selected examples. Based on the queried information, the classification model will be updated, which can improve the robustness of the model as much as possible at a lower cost.
% Based on the queried information, the perturbation levels of all training examples are adaptively adjusted, which are further used to update the classification model.

Formally, we define a triplet $(\mathbf{x}, y, \sigma)$ for each training example, which consists of the feature instance, the label and the instance-dependent perturbation level. Then the triplet dataset $T$ with $n$ examples is defined as follows:
\begin{equation}
T \coloneqq \{(\mathbf{x}_1,y_1,\sigma_1),(\mathbf{x}_2,y_2,\sigma_2),...,(\mathbf{x}_n,y_n,\sigma_n)\}.
\end{equation}

Next, we will discuss how to select the most useful examples from $T$ to query the perturbation level. As discussed before, neither excessive nor deficient perturbation is helpful to improve the model robustness. If an example falls into these cases, then its perturbation should be corrected to a proper level to conform with its intrinsic robustness. Therefore, given a triplet $(\mathbf{x},y,\sigma)$, the conformity $s(\sigma)$ of the perturbation to an example $\mathbf{x}$ can be defined as the perturbation level change before and after querying:
\begin{equation}\label{Eq:suitability}
s(\sigma) \coloneqq \sigma-\sigma_o,
\end{equation}
where $\sigma_o$ is the optimal perturbation level of $\mathbf{x}$, which corresponds to the maximum perturbation that the oracle can bear to identify the semantic contents of an image. Intuitively, the larger difference between the current perturbation level and the optimal perturbation level, the more helpful information it may gain with the correction. This motivates us to select the examples that are least conform with its intrinsic robustness.

However, we cannot get the optimal level $\sigma_o$ before active queries. That is why we have to find a surrogate of the conformity $s(\sigma)$.
Inspired by randomized smoothing~\cite{cohen2019certified}, we define the classification entropy to estimate the conformity of the perturbation level for an example. Specifically, for an example $\mathbf{x}$, we firstly generate $M$ noise instances with additive Gaussian noise $\mathcal{N}(0,\sigma^2 I)$. And then, the current classifier $F$ will predict the classes of these $M$ noise examples. Intuitively, if the $M$ predictions are highly consistent (with small entropy), then it implies that the example $\mathbf{x}$ has deficient perturbation. On the other hand, if the $M$ predictions are inconsistent (with large entropy), then it is likely that $\mathbf{x}$ received excessive perturbation currently, and thus its perturbation level may need correction from the oracle.

Formally, suppose that we have a classifier $F$ and an input $\mathbf{x}$, the probability of being classified as class $k$ under perturbations is $p_k \coloneqq \mathbb{P}(F(\mathbf{x}+\boldsymbol{\varepsilon})=k)$, where $\boldsymbol{\varepsilon} \sim \mathcal{P}(\sigma)$ and the noise distribution $\mathcal{P}(\sigma)$ can be any general noise distribution. Without loss of generality, we choose $\mathcal{P}(\sigma) = \mathcal{N}(0, \sigma^{2} I)$ as an example in this paper. Then the classification entropy can be defined as follows:
\begin{equation}\label{Eq:entropy}
H \coloneqq -\sum_{k=1}^{K} p_k \operatorname{ln}(p_k),
\end{equation}
where the probability $p_k$ is estimated using Monte Carlo sampling as discussed above.

\begin{figure}[t]
	\begin{center}
		\includegraphics[width=0.19\textwidth]{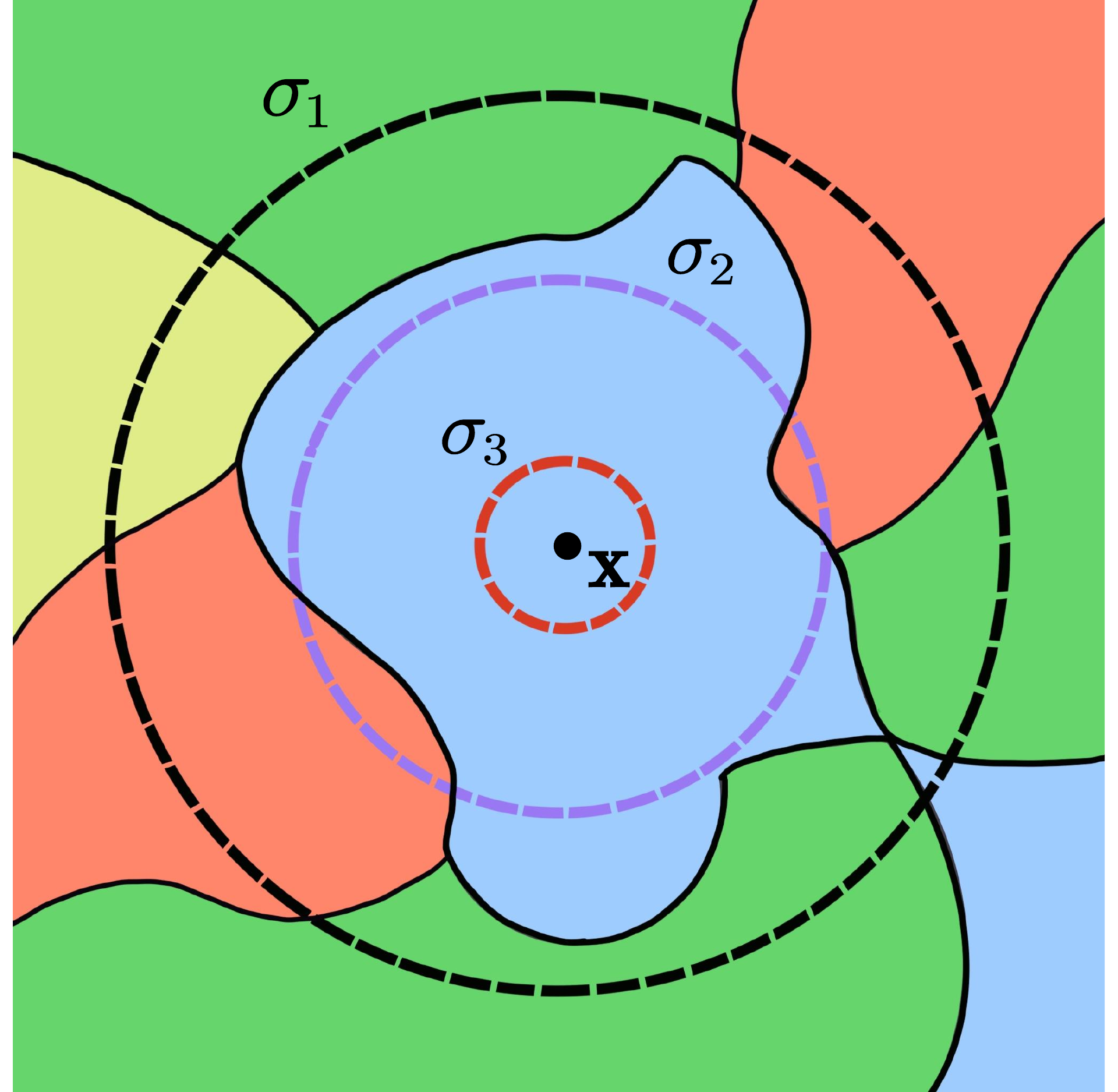}
		\includegraphics[width=0.27\textwidth]{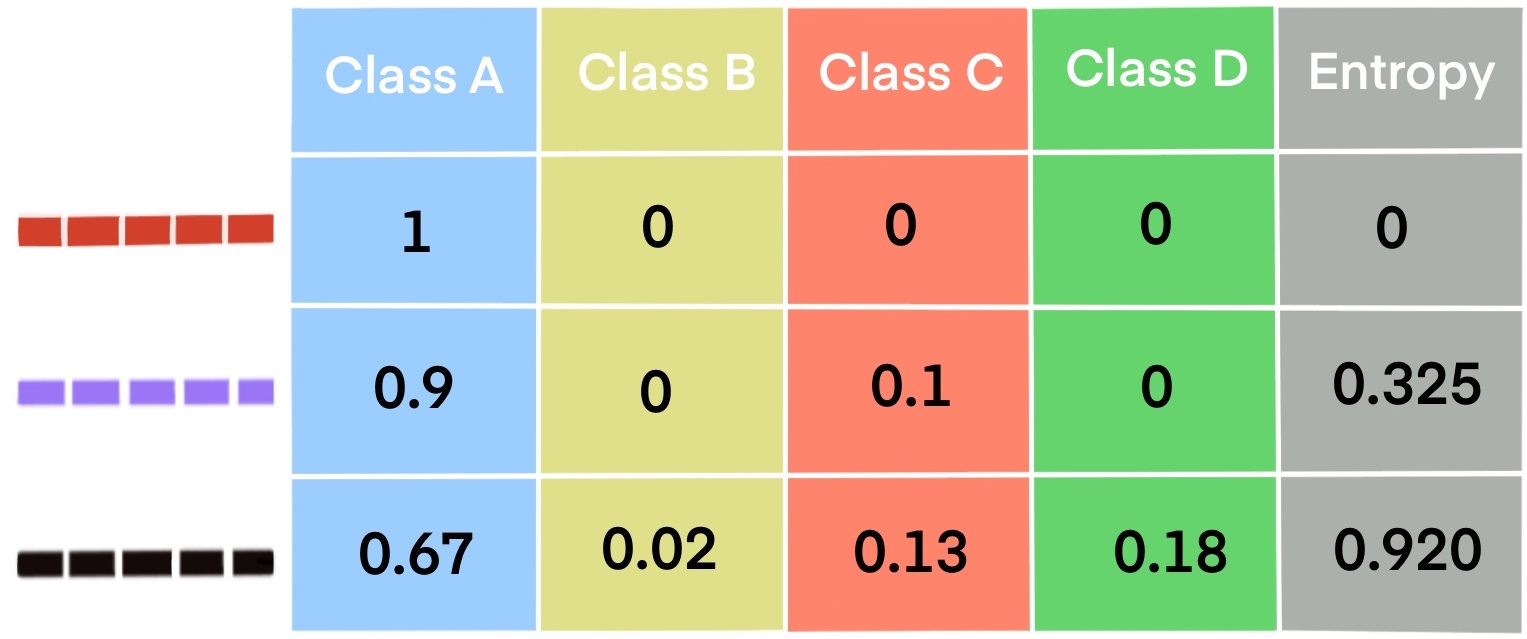}
	\end{center}
	\caption{The decision regions of the current classifier $F_\theta$ are drawn in different color, and the three circles respectively represent different perturbation levels $\sigma_i$ of the distributions $\mathcal{N}(\mathbf{x}, \sigma_i^2I)$, where $\sigma_1 > \sigma_2 > \sigma_3$. The corresponding class distribution and the entropy are shown in the table, which indicates that a larger perturbation leads to a larger entropy.}
	\label{fig:intuitionAndTheroy}
\end{figure}

Figure \ref{fig:intuitionAndTheroy} presents an example to show the relation between the perturbation level and classification entropy. It can be observed that with an excessive perturbation, the classifier (corresponding to the black circle) will produce uncertain predictions with a large entropy, while with a deficient perturbation, the classifier (corresponding to the red circle) will produce consistent predictions with a small entropy. Based on the classification entropy, we then select the two examples with least conformities (one with most excessive perturbation and one with most deficient perturbation) to query their correct perturbation levels from the oracle.

\begin{algorithm}[t]
	
	\caption{The AQPL algorithm}
	\label{alg:AQPL}
	\begin{algorithmic}[1]
		\STATE \textbf{Input:}
		\STATE \quad Query batch size $B$;
		\STATE \quad Triplet dataset $T$;
		\STATE \quad Pretrained model $F_\theta$;
		%		\STATE \textbf{Process:}
		%		\STATE \quad Initialize model parameter $\theta$ randomly.
		%		\STATE \quad Initialize perturbation level as a constant $\sigma_1=\sigma_2=...=c$;
		\STATE \textbf{Repeat:}
		%		\STATE \quad \textbf{Repeat:}
		%		\STATE \quad \quad Randomly select a batch $\{(\mathbf{x}_i,y_i,\sigma_i)\}_i^B$ from $T$.
		%		\STATE \quad \quad Randomly perturb 50\% of this batch with Gaussian noise.
		%		\STATE \quad \quad Update $\theta$ by minimizing Eq~\ref{EQ2}.
		%		\STATE \quad \textbf{until} convergence
		
		\STATE \quad Generate $M$ noise examples with additive Gaussian noise $\mathcal{N}(0,\sigma_i^2 I)$ for each example $\mathbf{x}_i$.
		\STATE \quad Caculate the classification entropy $H(\mathbf{x}_i)$ for each example $\mathbf{x}_i$.  
		
		\STATE \quad Select two batches of examples with maximum and minimum entropy.
		\STATE \quad Query the most acceptable perturbation level $\sigma^*$ for selected examples from oracle.
		
%		\STATE \quad Select two examples $\mathbf{x}_j$ and $\mathbf{x}_k$ with maximum and minimum entropy.
%		\STATE \quad Query the most acceptable perturbation level $\sigma_j^*$ and $\sigma_k^*$ from oracle.
%		\STATE \quad Adjust the perturbation levels $\sigma_j^*$ for top $\frac{B}{2}$ examples with the largest entropy.
%		\STATE \quad Adjust the perturbation levels $\sigma_k^*$ for top $\frac{B}{2}$ examples with the smallest entropy.
		
		\STATE \quad Update the triplet dataset $T$ and update $\theta$ by minimizing Eq~\ref{EQ2}.
		\STATE \textbf{until} query budget or expected performance reached.
		\STATE Output the learned model $F_\theta$.
		
	\end{algorithmic}
\end{algorithm}

Next we discuss how to let the oracle decide the proper perturbation level for the selected examples. Intuitively, if the corrupted image is difficult for a human annotator to identify its semantic content, then it is likely that the image is suffering from excessive perturbation. For the selected example $\mathbf{x}^*$, we generate a series of noise images from the clean instance by varying the perturbation level from the minimum $\sigma_{min}$ to the maximum $\sigma_{max}$ with interval of $\alpha$. Among which, the oracle is asked to choose the image that is at the threshold of identifying its semantic content. Then the corresponding perturbation level of this image is annotated as the optimal perturbation level for the queried example $\mathbf{x}^*$. This annotation process is illustrated in the dashed rectangle in Figure \ref{fig:AQPL}. Among the noise images generated from the selected example, the black marker indicates the one generated with the current perturbation level, while the red marker indicates the optimal perturbation level annotated by the oracle.

After the querying, the triplet with corrected perturbation level is added into the training set for updating the model.
Moreover, to improve the efficiency of learning, we also query the most acceptable perturbation level for two mini batches of examples with maximum and minimum entropy. 
At last, we update $\theta$ by minimizing Eq~\ref{EQ2} until query budget or expected performance reached. 
Note that to maintain high accuracy on clean data, we only perturb 50\% of the training data with Gaussian noise to train the model within each batch, which follows the same settings in~\cite{rusak2020increasing}.

The process of the approach is summarized in Algorithm~\ref{alg:AQPL}. Firstly, triplet dataset $T$, query batch size $B$ and pretrained model $F_\theta$ are given. Then for each example $\mathbf{x}_i$, we generate $K$ noise examples with additive Gaussian noise $\mathcal{N}(0,\sigma_i^2 I)$ and caculate its classification entropy $H(\mathbf{x}_i)$. 
We select two batches of examples with maximum and minimum entropy and ask for annotations of perturbation level $\sigma^*$. After that, we update the triplet dataset $T$ and update $\theta$ by minimizing Eq~\ref{EQ2} until query budget or expected performance reached.

\subsection{Theoretical Analysis}

In this subsection, we present theoretical analysis to show the rationality of the proposed classification entropy as a surrogate of the conformity for a simple linear case. Although it doesn't extend easily to the non-linear deep learning based classification, this analysis gives some insights into the behavior of the proposed surrogate, and how this strategy successfully reduces the query number for robust learning.

For the definition of conformity in Eq \ref{Eq:suitability}, if we know the optimal perturbation level $\sigma_o$, the triplet with largest or smallest value of conformity $s(\sigma)$  will be selected by the proposed algorithm. In other words, the perturbation level that changes the most before and after querying is of our interest. With the proposed method to estimate $s(\sigma)$ by the classification entropy $H$ in Eq \ref{Eq:entropy}, we can get the following theorems.

\begin{restatable}[]{theorem}{thmA}
	\label{theorem:thmA}
	Consider the case of one-layer feed-forward network for binary classification $F(\mathbf{x}) = \operatorname{sign}(f(\mathbf{x}))$ and $f(\mathbf{x}) = \mathbf{w}^T \mathbf{x} + b$, where $\mathbf{w} \in \mathbb{R}^d$ and $b \in \mathbb{R}$.
	For any $\mathbf{x} \in \{\mathbf{x}: f(\mathbf{x}) \neq 0\}$, suppose that $\mathcal{P}(\sigma) = \mathcal{N}(0, \sigma^{2} I)$ and its current perturbation level $\sigma \in (0, \infty)$, then we have
	\begin{align}
	\sigma \propto H.
	\end{align}
\end{restatable}

The proof can be found in the supplementary material. Further, if there is an oracle classifier $F_o(\mathbf{x})$, then for the optimal perturbation level $\sigma_o$ we have
\begin{align}
\mathbb{P}(F_o(\mathbf{x}+\boldsymbol{\varepsilon}_o) = c) = \tau, 
\end{align}
where $\boldsymbol{\varepsilon}_o  \sim \mathcal{P}(\sigma_o)$ and $\tau$ is some sufficient large value (e.g. $99.73\%$ for the empirical rule~\cite{pukelsheim1994three}). Then for any fixed $\sigma_o$ and $\tau$, we have following theorem.

\begin{restatable}[]{theorem}{thmB}
	\label{theorem:thmB}
	Consider the case of one-layer feed-forward network for binary classification $F(\mathbf{x}) = \operatorname{sign}(f(\mathbf{x}))$ and $f(\mathbf{x}) = \mathbf{w}^T \mathbf{x} + b$, where $\mathbf{w} \in \mathbb{R}^d$ and $b \in \mathbb{R}$.
	%	Given any $\mathbf{x} \in \{\mathbf{x}: f(\mathbf{x}) \neq 0\}$ and any current perturbation level $\sigma \in (0, \infty)$, suppose that its optimal perturbation level is $\sigma_o$ for the oracle classifier $F_o(\mathbf{x})=\operatorname{sign}(\mathbf{w}_o^T\mathbf{x}+b_o)$. If $\mathbf{w}^T \mathbf{w}_o \neq 0$, then we have
	For every $\mathbf{x} \in \{\mathbf{x}: f(\mathbf{x}) \neq 0\}$ and its corresponding optimal perturbation level $\sigma_o$ for the oracle classifier $F_o(\mathbf{x})=\operatorname{sign}(\mathbf{w}_o^T\mathbf{x}+b_o)$, suppose that $\mathcal{P}(\sigma) = \mathcal{N}(0, \sigma^{2} I)$. If $\mathbf{w}^T \mathbf{w}_o \neq 0$, then we have
	\begin{align}
	\sigma_o \propto -H.
	\end{align}
\end{restatable}
The proof can be found in the supplementary material. Then we can further get the relation between the conformity $s(\sigma) \coloneqq \sigma-\sigma_o$ and the classification entropy $H$ with the following corollary.

\begin{corollary}
	\label{corollary:cor1}
	Consider the case of linear binary classifier $F(\mathbf{x}) = \operatorname{sign}(f(\mathbf{x}))$ and $f(\mathbf{x}) = \mathbf{w}^T \mathbf{x} + b$, where $\mathbf{w} \in \mathbb{R}^d$ and $b \in \mathbb{R}$.
	For every $\mathbf{x} \in \{\mathbf{x}: f(\mathbf{x}) \neq 0\}$ and its corresponding optimal perturbation level $\sigma_o$ for the oracle classifier $F_o(\mathbf{x})=\operatorname{sign}(\mathbf{w}_o^T\mathbf{x}+b_o)$, suppose that $\mathcal{P}(\sigma) = \mathcal{N}(0, \sigma^{2} I)$ and its current perturbation level $\sigma \in (0, \infty)$. If $\mathbf{w}^T \mathbf{w}_o \neq 0$, then we have
	\begin{align}
	\sigma - \sigma_o \propto H.
	\end{align}
\end{corollary}

With Corollary~\ref{corollary:cor1}, it can be observed that the conformity of a perturbation level to an example is directly proportional to the classification entropy. In other words, based on the classification entropy $H$ of example $\mathbf{x}$, we can effectively select the examples which are most helpful for improving the robustness after correcting their perturbation levels.

\begin{figure*}[!t]
	\centering
	
	\begin{subfigure}{0.3\linewidth}
		\centering
		\label{fig.Mnist.Gaussian}
		\includegraphics[width=1\textwidth]{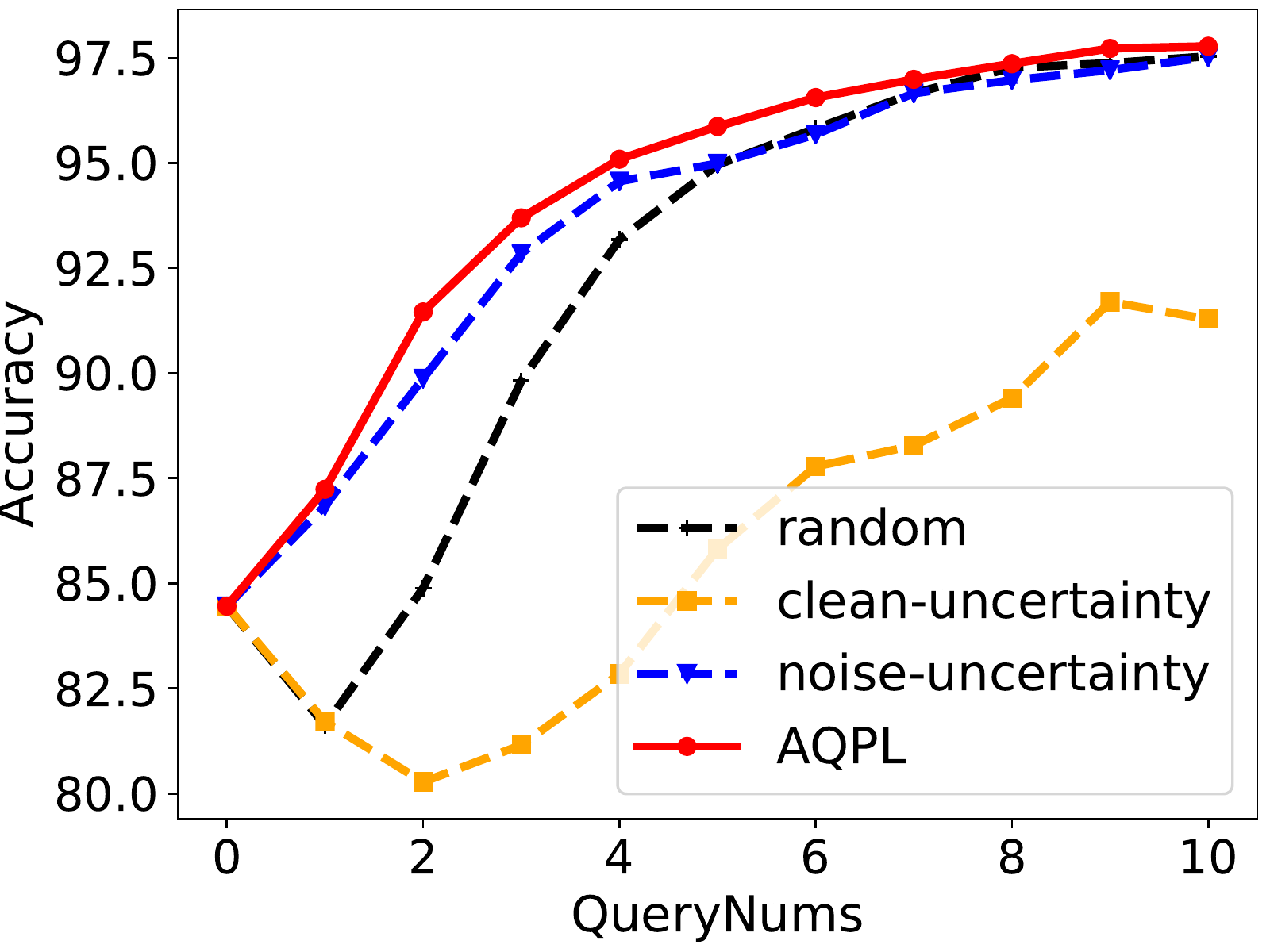}\\
		\caption{MNIST-C}
	\end{subfigure}
	\begin{subfigure}{0.3\linewidth}
		\centering
		\label{fig.Cifar10.Gaussian}
		\includegraphics[width=1\textwidth]{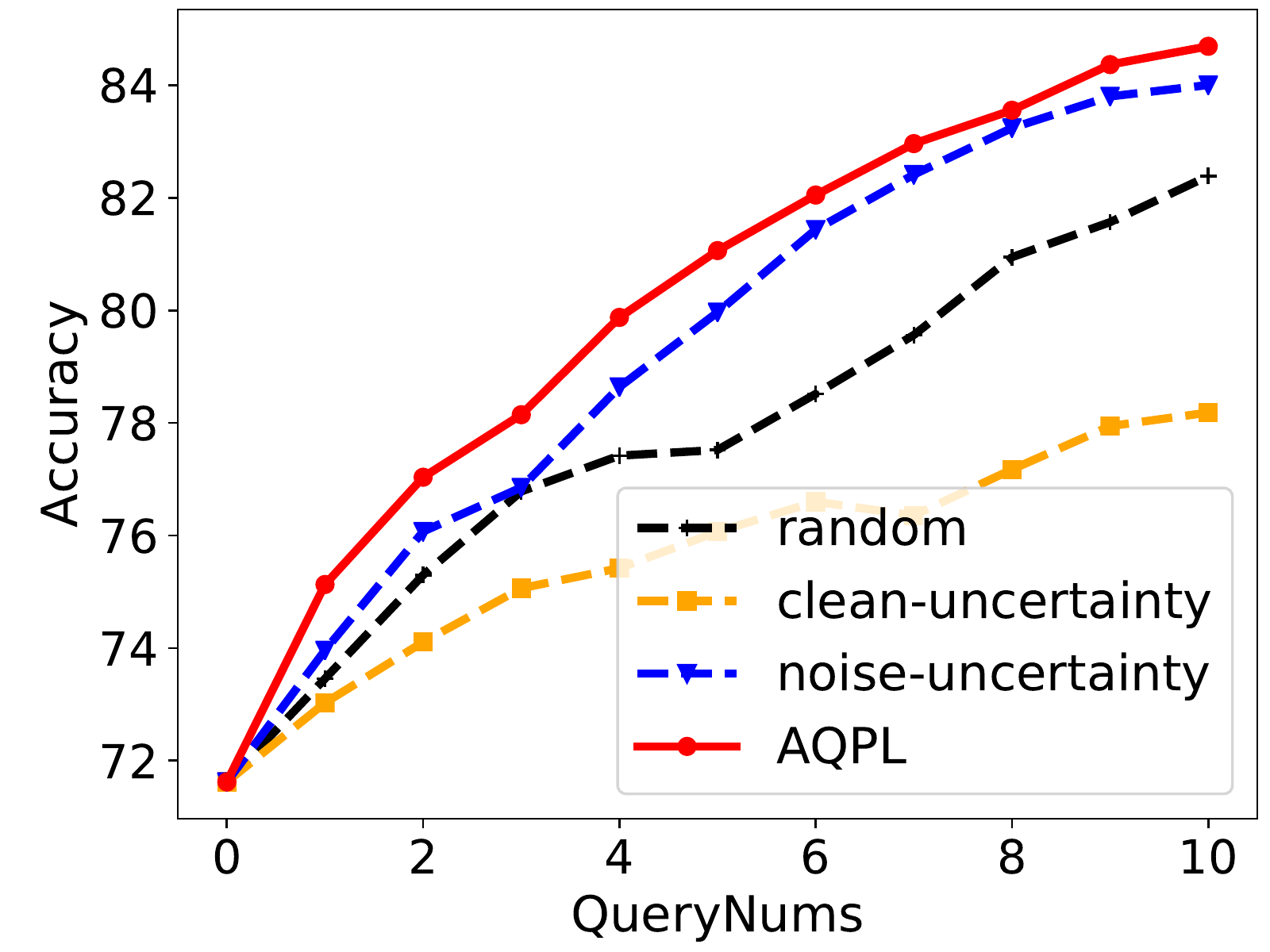}\\
		\caption{CIFAR10-C}
	\end{subfigure}
	\begin{subfigure}{0.3\linewidth}
		\centering
		\label{fig.TinyImagenet.Gaussian}
		\includegraphics[width=1\textwidth]{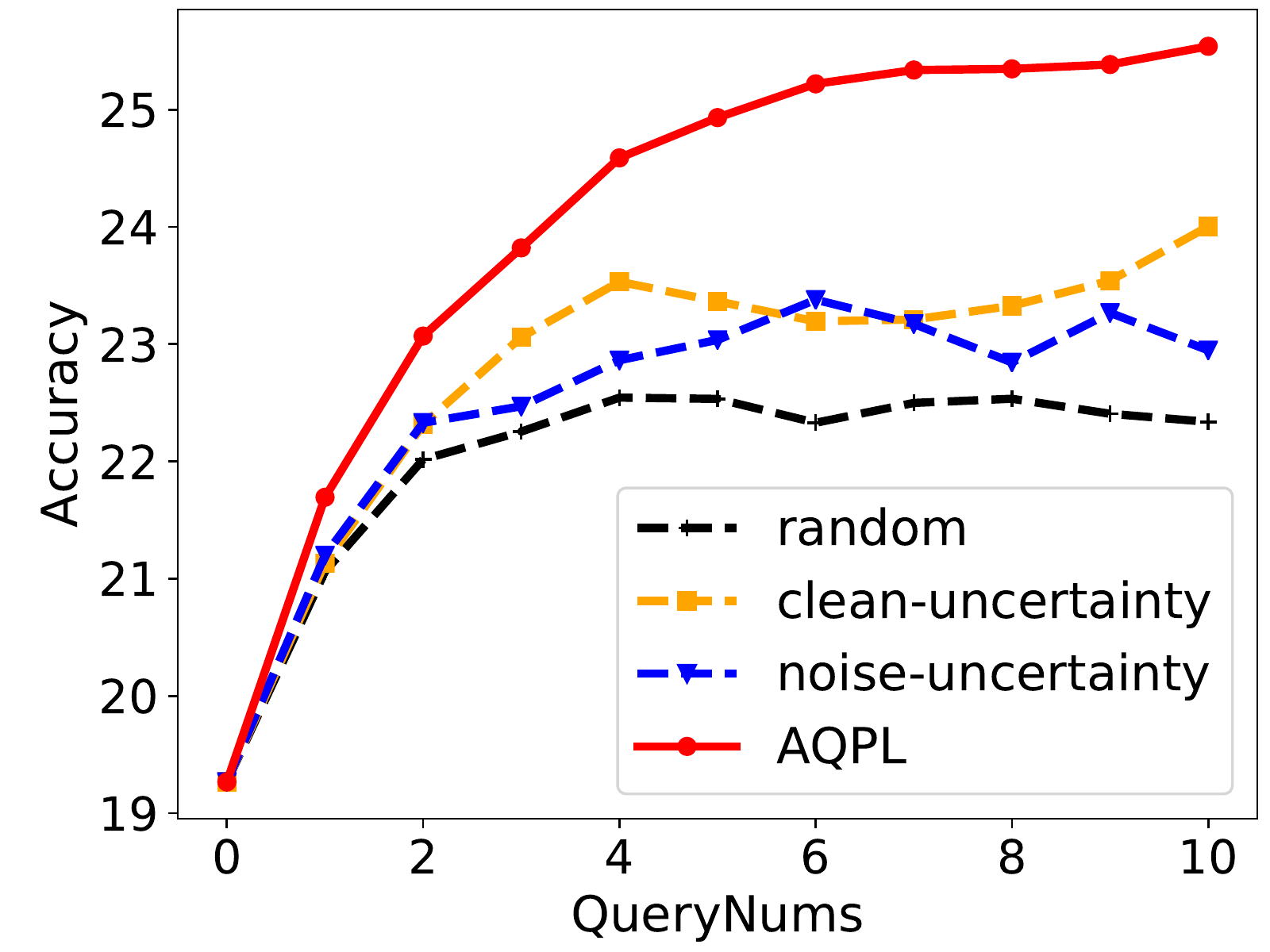}\\
		\caption{Tiny-Imagenet-C}
	\end{subfigure}
	
	\caption{Performance comparison of different methods towards Gaussian noise in corruption datasets.}
	\label{fig.comp.Gaussian}
\end{figure*}

\begin{figure*}[!t]
	\centering
	
	\begin{subfigure}{0.3\linewidth}
		\centering
		\label{fig.Mnist.All}
		\includegraphics[width=1\textwidth]{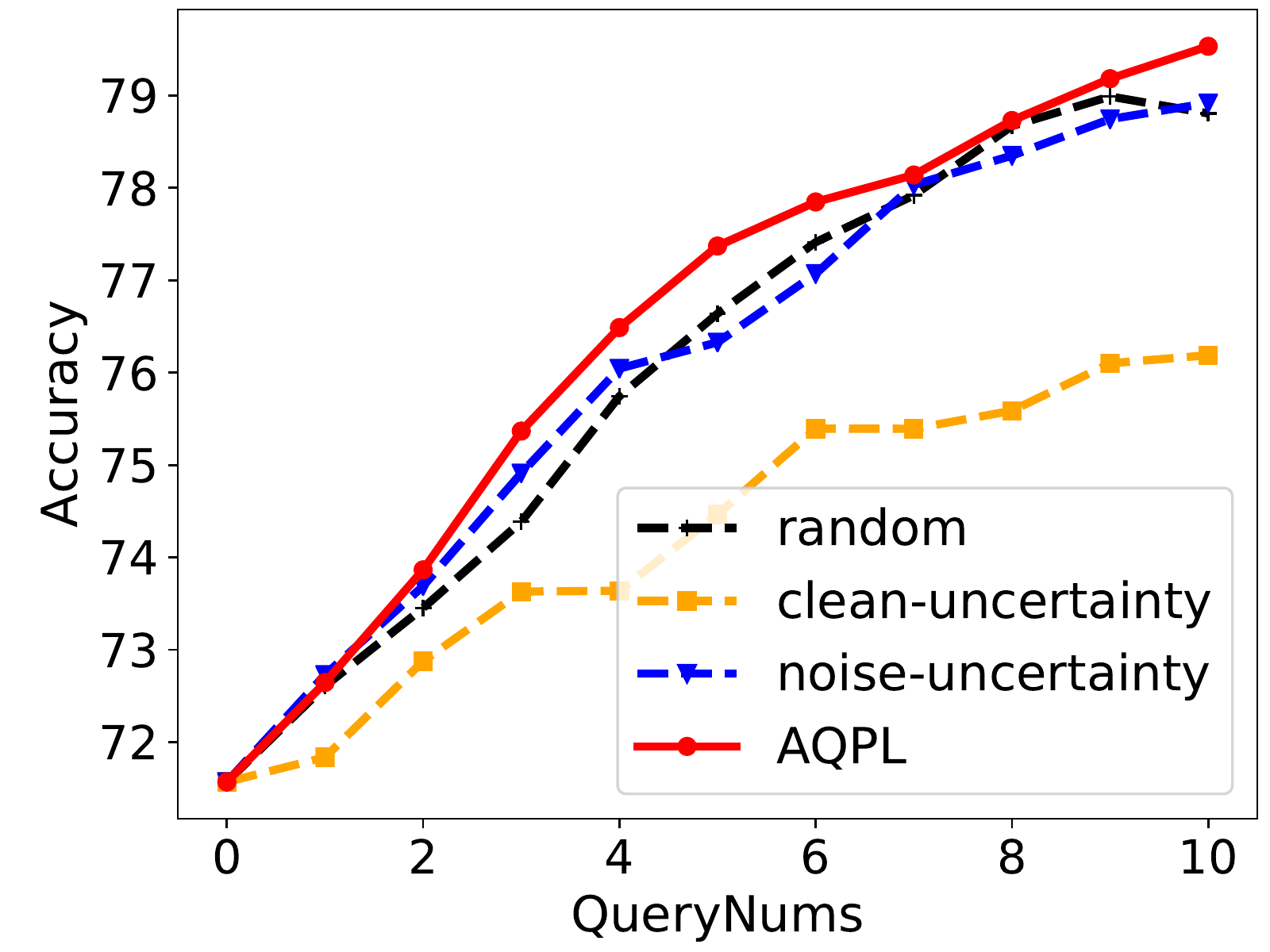}\\
		\caption{MNIST-C}
	\end{subfigure}
	\begin{subfigure}{0.3\linewidth}
		\centering
		\label{fig.Cifar10.All}
		\includegraphics[width=1\textwidth]{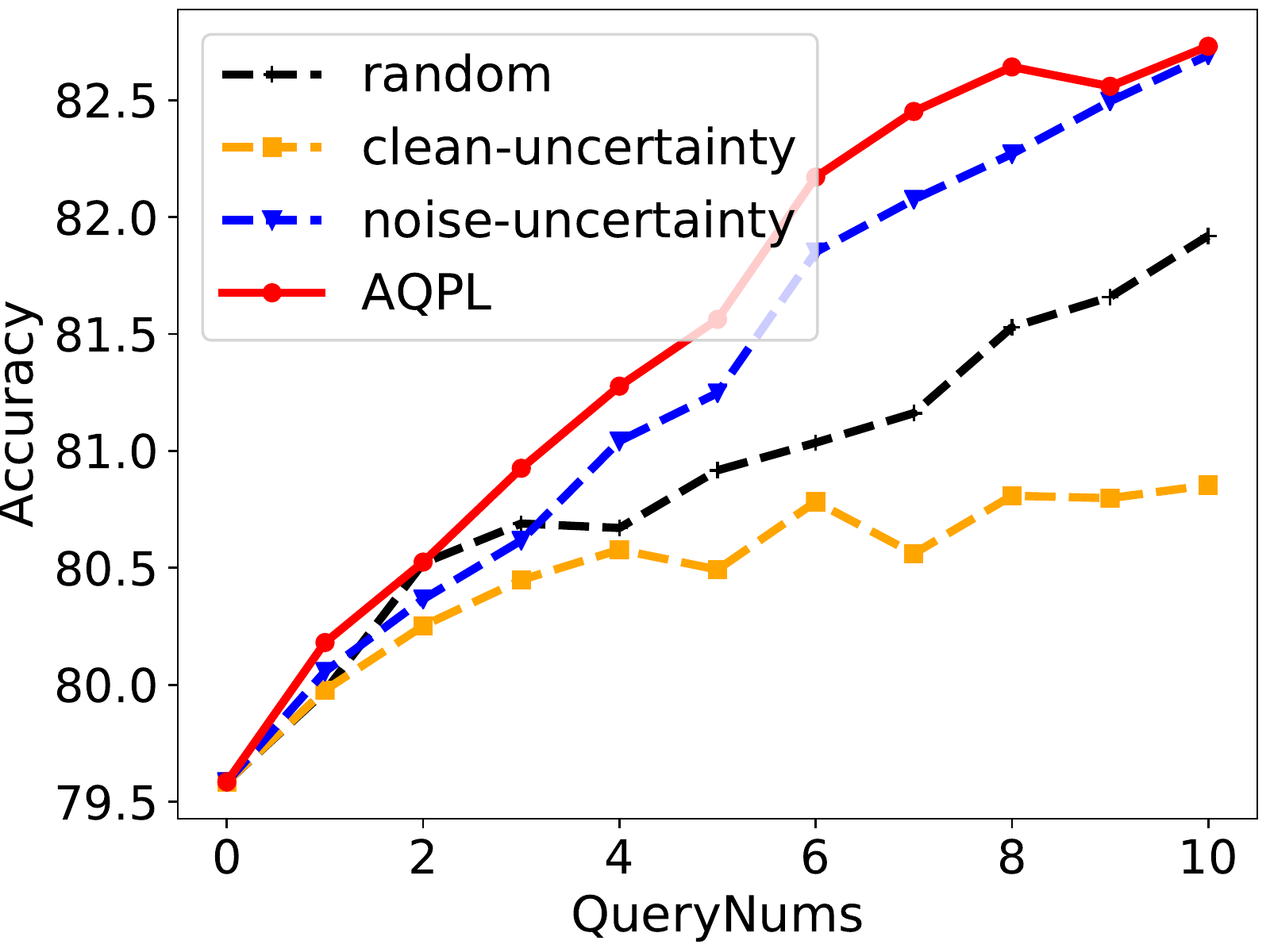}\\
		\caption{CIFAR10-C}
	\end{subfigure}
	\begin{subfigure}{0.3\linewidth}
		\centering
		\label{fig.TinyImagenet.All}
		\includegraphics[width=1\textwidth]{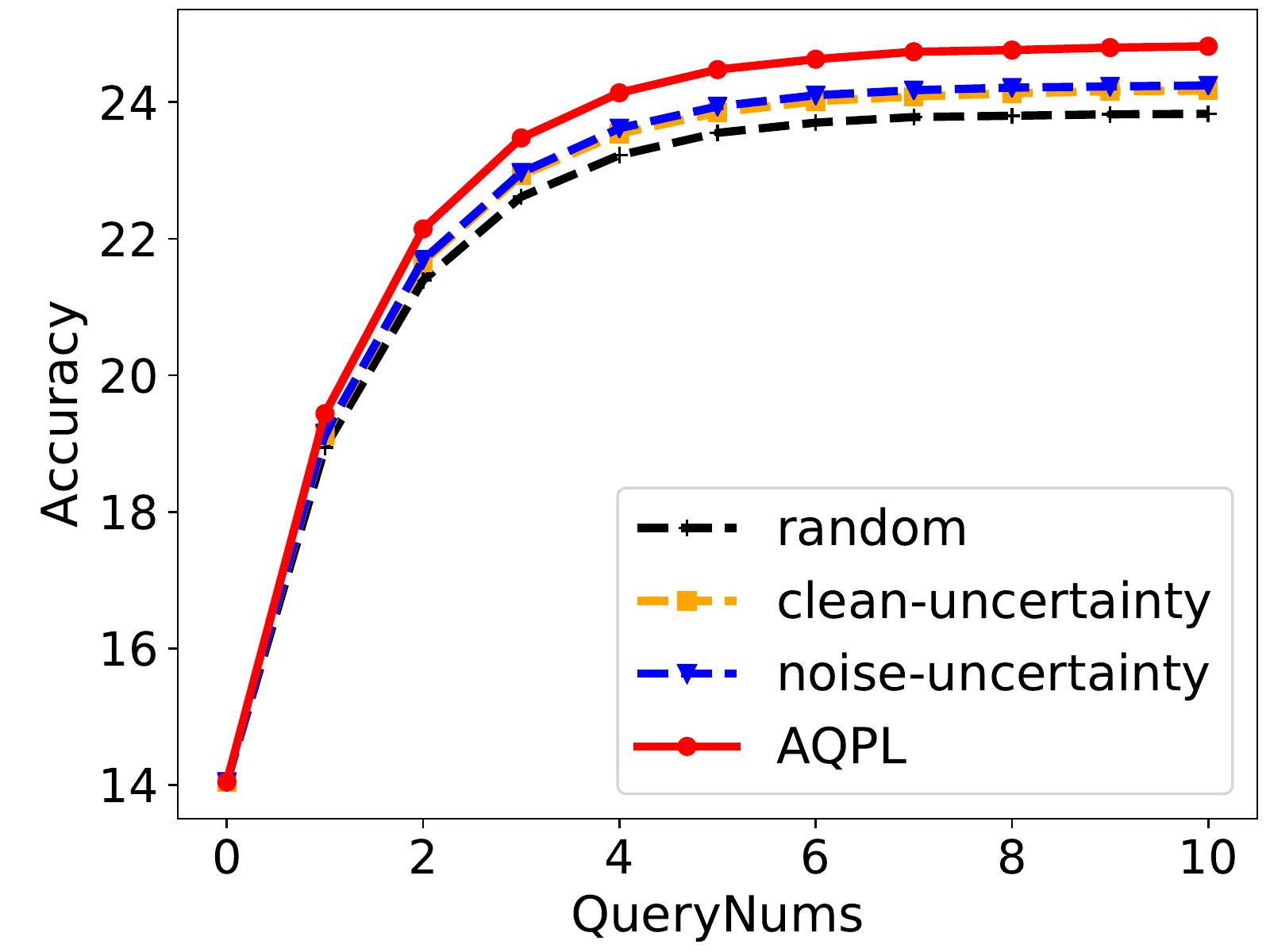}\\
		\caption{Tiny-Imagenet-C}
	\end{subfigure}
	
	\caption{Average performance comparison of different methods towards 15 types of noise in corruption datasets.}
	\label{fig.comp.all}
\end{figure*}

\section{Experiments}

\subsection{Settings}
To validate the effectiveness of the proposed approach, we perform experiments on six datasets. Specifically, we train the model on MNIST~\cite{lecun1998gradient}, CIFAR10~\cite{krizhevsky2009learning} and Tiny-Imagenet~\cite{yao2015tiny}, and test on MNIST-C~\cite{mu2019mnist}, CIFAR10-C~\cite{hendrycks2019robustness}, Tiny-Imagenet-C~\cite{hendrycks2019robustness}. 
We employ ResNet18 model~\cite{he2016deep} as the base model to implement our approach as well as other compared methods. 

We respectively examine the effectiveness of our approach with regard to the active sampling strategy and the querying method. To validate the effectiveness of the sampling strategy, we compare the following methods in the experiments:
\emph{i)} \textbf{Random}: it selects examples at random.
\emph{ii)} \textbf{Clean-uncertainty}: it selects the examples with largest uncertainty of clean example predictions~\cite{lewis1994sequential}.
\emph{iii)} \textbf{Noise-uncertainty}: a reasonable extension of the last strategy. It selects the examples with largest expected uncertainty of noise example preditions. Specifically, it uses the current perturbation level and clean example to generate $M$ noise examples and selects the examples with the largest average uncertainty of these noise examples predictions.
\emph{iv)} \textbf{AQPL (ours)}: the propose approach. It selects the examples with most unsuitability of noise to examples.

Moreover, to validate the effectiveness of the querying method, the following methods are compared:
\emph{i)} \textbf{Standard}: the model are trained only on clean datasets.
\emph{ii)} \textbf{GNT}~\cite{rusak2020increasing}: it uses a fixed perturbation level to perturb 50\% of the training data with  Gaussian noise within each batch, and trains the model with clean data and noise data.
\emph{iii)} \textbf{CAT}~\cite{cheng2020cat}: it adaptively customizes the perturbation level according to whether the model has capacity to robustly classify the example.
\emph{iv)} \textbf{AQPL-GNT (ours)}: the proposed approach. It corrects perturbation levels by interacting human experts, which based on the training model of GNT.
\emph{v)} \textbf{AQPL-CAT (ours)}: the proposed approach. It corrects perturbation levels by interacting human experts, which based on the training model of CAT.

For all active learning methods, $M$ is set to 50, and we fix the query batch size $B$ to 100 on CIFAR10 and MNIST, and 500 on Tiny-Imagenet at each active querying iteration. In annotation process, the parameters $\sigma_{min}$, $\sigma_{max}$ and $\alpha$ are respectively set to 0, 0.9 and 0.01. More hyper-parameters and experimental details can be found in the supplementary material.

\begin{table*}[!t]		
	\centering
	\begin{tabular}{c|c|ccccc}
		\hline
		\multirow{2}{*}{Dataset} & \multirow{2}{*}{Type}
		&\multicolumn{5}{c}{Method}\\
		\cline{3-7}
% 		&&Standard & GNT~\cite{rusak2020increasing} & CAT~\cite{cheng2020cat} & AQPL-GNT & AQPL-CAT\\
		&&Standard & GNT & CAT & AQPL-GNT & AQPL-CAT\\
		\hline
		\multirow{3}{*}{MNIST-C}
		& Clean & $\bm{99.29\%}$ & $97.32\%$ & $98.43\%$ & $99.21\%$ & $99.23\%$ \\
		& Gaussian & $16.06\%$ & $84.46\%$ & $\bm{98.14\%}$ & $96.31\%$ & $97.90\%$ \\
		& All & $65.34\%$ & $71.57\%$ & $80.11\%$ & $78.78\%$ & $\bm{80.42\%}$ \\
		\hline
		
		\multirow{3}{*}{CIFAR10-C}
		& Clean & $\bm{95.05\%}$ & $94.87\%$ & $86.42\%$ & $94.83\%$ & $94.75\%$ \\
		& Gaussian & $43.23\%$ & $71.62\%$ & $82.78\%$ & $82.19\%$ & $\bm{86.69\%}$ \\
		& All & $74.24\%$ & $79.59\%$ & $71.15\%$ & $82.02\%$ & $\bm{83.33\%}$ \\
		\hline
		
		\multirow{3}{*}{Tiny-Imagenet-C}
		& Clean & $\bm{57.84\%}$ & $56.14\%$ & $48.62\%$ & $56.60\%$ & $55.51\%$ \\
		& Gaussian & $19.27\%$ & $21.90\%$ & $27.98\%$ & $25.12\%$ & $\bm{31.72\%}$ \\
		& All & $9.99\%$ & $14.04\%$ & $23.77\%$ & $24.82\%$ & $\bm{27.19\%}$ \\
		
		\hline
	\end{tabular}
	\caption{The Top-1 accuracy of different methods on different corruption datasets.}
	\label{table:comparison}
\end{table*}

% \begin{table*}[!t]		
% 	\centering
% 	\begin{tabular}{c|c|lllll}
% 		\hline
% 		\multirow{2}{*}{Dataset} & \multirow{2}{*}{Type}
% 		&\multicolumn{5}{c}{Method}\\
% 		\cline{3-7}
% % 		&&Standard & GNT~\cite{rusak2020increasing} & CAT~\cite{cheng2020cat} & AQPL-GNT & AQPL-CAT\\
% 		&&Standard & GNT & CAT & AQPL-GNT & AQPL-CAT\\
% 		\hline
% 		\multirow{3}{*}{MNIST-C}
% 		& Clean & $\bm{99.29\%}$ & $97.32\%$ & $98.43\%$ & $99.21\% (+1.89\%)$ & $99.23\% (+0.80\%)$ \\
% 		& Gaussian & $16.06\%$ & $84.46\%$ & $\bm{98.14\%}$ & $96.31\% (+11.85\%)$ & $97.90\% (-0.24\%)$ \\
% 		& All & $65.34\%$ & $71.57\%$ & $80.11\%$ & $78.78\% (+7.21\%)$ & $\bm{80.42\%} (+0.31\%)$ \\
% 		\hline
		
% 		\multirow{3}{*}{CIFAR10-C}
% 		& Clean & $\bm{95.05\%}$ & $94.87\%$ & $86.42\%$ & $94.83\% (-0.04\%)$ & $94.75\% (+8.33\%)$ \\
% 		& Gaussian & $43.23\%$ & $71.62\%$ & $82.78\%$ & $82.19\% (+10.57\%)$ & $\bm{86.69\%} (+3.91\%)$ \\
% 		& All & $74.24\%$ & $79.59\%$ & $71.15\%$ & $82.02\% (+2.43\%)$ & $\bm{83.33\%} (+12.18\%)$ \\
% 		\hline
		
% 		\multirow{3}{*}{Tiny-Imagenet-C}
% 		& Clean & $\bm{57.00\%}$ & $56.00\%$ & $48.00\%$ & $56.00\% (+0.00\%)$ & $55.00\% (+7.00\%)$ \\
% 		& Gaussian & $19.27\%$ & $21.90\%$ & $27.98\%$ & $25.12\% (+3.22\%)$ & $\bm{31.72\%} (+3.74\%)$ \\
% 		& All & $9.99\%$ & $14.04\%$ & $23.77\%$ & $24.82\% (+10.78\%)$ & $\bm{27.19\%} (+3.42\%)$ \\
		
% 		\hline
% 	\end{tabular}
% 	\caption{The Top-1 accuracy of different methods on different corruption datasets.}
% 	\label{table:comparison}
% \end{table*}

\subsection{Performance comparison}
We plot the accuracy curves of the proposed AQPL approach and compared methods with the number of queries increasing. The results with Gaussian noise are shown in Figure~\ref{fig.comp.Gaussian}. Because of the space limitation, we present the detailed results with other types of noises in the supplementary material, and show the average results of 15 types of noise in Figure~\ref{fig.comp.all}. The term "QueryNums" in all figures refers to the epoch of interactions with the oracle, and two batches of examples are queried from the oracle at each epoch. It is worthy to note that when comparing with other methods, we use the same base model and query batches of the same size to update the base model for fair comparison. In addition, when the query number is $0$, all perturbation levels have not been updated, and the initial value is the performance of GNT.

From Figure~\ref{fig.comp.Gaussian} and~\ref{fig.comp.all}, we can observe that the proposed AQPL approach outperforms the other methods in all cases. AQPL can achieve higher accuracy with fewer queries on corruption datasets. The random method, which selects examples at random, can improve the model robustness by querying perturbation levels. This phenomenon implies that it is a reasonable way to improve model robustness by correcting perturbation levels with queried information. It is observed that the clean-uncertainty method performs poorly in most cases. One possible reason is that if the model is much uncertain about the clean example itself, then changing the perturbation level will not improve the model robustness. The noise-uncertainty method can always achieve suboptimal performance because noise examples with high uncertainty often need to adjust the perturbation level. The results in Figure~\ref{fig.comp.Gaussian} and~\ref{fig.comp.all} are consistent in general, validating that the proposed  AQPL can effectively improve the model robustness with fewer queries against different types of noises.

To further validate the effectiveness of the querying method, we also show the Top-1 accuracy achieved by different methods on different corruption datasets in Table~\ref{table:comparison}. It is worthy to note that the proposed AQPL-GNT and AQPL-CAT methods respectively use GNT~\cite{rusak2020increasing} and CAT~\cite{cheng2020cat} as the based model, and the mean results over 10 queries are recorded. First of all, the standard method can always achieve the best performance on the clean test set, while performs poorly on corruption datasets. When comparing with the method that only trains on the clean datasets, the GNT method, which trains with Gaussian noise with a fixed perturbation level, significantly improves the model robustness against various noises. The CAT method has higher performance on corruption datasets than GNT by adaptively customizing the perturbation levels of examples, which implies that it is important to adaptively adjust the perturbation levels for different examples in the training process. Moreover, by allowing to query the ground-truth information on the perturbation level, the proposed approaches AQPL-GNT and AQPL-CAT can further improve the performances of GNT and CAT respectively. Most importantly, it can be observed that the proposed approach AQPL-CAT outperforms the other methods in most cases with regard to both Gaussian noise and the other 15 types of noise. Note that, when comparing with the method CAT that also adjusts perturbation level according to whether the current model has the capacity to robustly classify the examples, the AQPL-CAT can still achieve better performance. On one hand, the supervised information provided by the oracle is more reliable. On the other hand, human experts correct perturbation levels more efficiently and directly.

In summary, these results consistently demonstrate that the proposed AQPL approach can effectively improve the model robustness by actively querying the correct perturbation level from the oracle, while the sampling strategy can efficiently select the most useful examples to reduce the querying cost.

\subsection{Discussion}

Similar to many existing studies, the experiments are performed on image datasets in this paper. The results show that, by actively querying the supervised information about the perturbation level, model robustness against corruption perturbations on image classification tasks can be improved efficiently. In principle, the proposed method can be applied to any type of data. One challenge is that it could be difficult for human annotators to select a proper perturbation level for non-visual data. If the non-visual data can be easily visualized, such as VisArtico \cite{ouni2012visartico} for articulatory data, the method is still applicable. It would be an interesting future work to design feasible interfaces for annotators to decide the perturbation level for non-visual data. 

In this paper, we focus on the corruption perturbations both in our theoretical and experimental analysis. We believe that corruption perturbations commonly occur in real tasks. On the other hand, it would be interesting to extend the study for improving adversarial robustness. Actually, the average-case robustness under a specific noise distribution could bring non-negligible adversarial robustness \cite{wong2020learning}. More importantly, the optimal perturbation level for a clean example considered in this paper, essentially, represents an adversarial (worst-case) noise distribution on the example with regard to the oracle.

\section{Conclusion}

In this work, we propose a novel active learning framework to improve the model robustness by querying the conform perturbation levels. On one hand, instead of assuming a fixed noise for the whole training set, the perturbation levels are adjusted adaptively for different examples during the training process. On the other hand, by estimating the conformity with classification entropy, the most useful examples are actively selected to achieve effective learning with lower annotation cost. Both theoretical and empirical results validate the effectiveness of the proposed approach. In the future, we plan to extend the framework to handle adversarial perturbations.

\section{ Acknowledgments}

This research was supported by the National Key R\&D Program of China (2020AAA0107000), NSFC (62076128) and the China University S\&T Innovation Plan Guided by the Ministry of Education.

\bibliography{aaai21}
\end{document}